\documentclass{article}
\usepackage{graphicx}
\usepackage[table,xcdraw]{xcolor}
\usepackage{multirow}
\usepackage{amsmath}
\usepackage{amssymb}
\usepackage{bm}

\usepackage{booktabs}
\usepackage{pifont}
\PassOptionsToPackage{numbers, compress}{natbib}

\usepackage[preprint]{neurips_2026}

\usepackage[utf8]{inputenc} 
\usepackage[T1]{fontenc}    
\usepackage{hyperref}       
\usepackage{url}            
\usepackage{booktabs}       
\usepackage{amsfonts}       
\usepackage{nicefrac}       
\usepackage{microtype}      
\usepackage{xcolor}         
\usepackage{wrapfig}
\usepackage{booktabs}
\usepackage{capt-of}
\usepackage{needspace}
\usepackage{multirow}
\usepackage[table]{xcolor}
\definecolor{myorange}{RGB}{218, 93, 0}
\definecolor{myblue}{RGB}{0, 35, 102}
\newcommand{\impor}[1]{\textcolor{myorange}{\textbf{#1}}}
\newcommand{\drop}[1]{\textcolor{myorange}{\textbf{$\downarrow$ #1}}}
\title{Agree to Disagree: Multimodal Autonomous Negotiation and Calibration for Entity Representation Learning}

\author{%
  \textbf{Chenyi Xiong}$^{1}$,
  \textbf{Yan Zhang}$^{1,2}$,
  \textbf{Yueluan Huang}$^{1}$,
  \textbf{Lu Wang}$^{1}$,\\
  \textbf{Kui Xiao}$^{1,2}$,
  \textbf{Wenxin Huang}$^{1,2}$,
  \textbf{Zhifei Li}$^{1,2,*}$\\
  $^{1}$School of Computer Science, Hubei University\\
  $^{2}$Hubei Key Laboratory of Big Data Intelligent Analysis and Application, Hubei University\\
  \texttt{\{xiongchenyi, yueluan, luwang\}@stu.hubu.edu.cn}\\
  \texttt{wenxinhuang\_wh@163.com,\{zhangyan,xiaokui,zhifei1993\}@hubu.edu.cn}
}

\begin{document}
\begingroup
\renewcommand{\thefootnote}{*}
\footnotetext{Corresponding author.}
\endgroup

\maketitle

\begin{abstract}
Learning high-quality multimodal entity representations is essential for advancing reasoning tasks such as multimodal knowledge graph completion (MMKGC). However, rigid alignment in existing fusion strategies can bias representations toward a dominant anchor modality, weakening fine-grained complementary cues and propagating noisy signals into relational reasoning. To address these limitations, we propose \textbf{MANoR}, a \textbf{M}ultimodal \textbf{A}utonomous \textbf{N}eg\textbf{o}tiation framework for calibrated \textbf{R}epresentation learning. MANoR treats modalities as autonomous semantic agents that negotiate before fusion. Specifically, MANoR preserves modality-specific semantics through centroid-guided subspace contraction, which reduces unstable centroid-relative dispersion while retaining entity-level residual semantics. On this stabilized basis, negotiated attention enables selective cross-modal interaction without collapsing modality boundaries. As modalities can still differ in reliability after interaction, uncertainty calibrated fusion estimates dimension-level aleatoric uncertainty and modality-level reliability to compute relation-aware fusion weights. In this way, MANoR couples modality autonomy, selective interaction, and calibrated integration within a unified representation learning framework. Experiments on MKG-W, MKG-Y, TIVA, and KVC16K show that MANoR consistently outperforms strong MMKGC baselines, achieving a relative Hits@1 gain of up to 17.72\% on KVC16K. Our code can be found at \url{https://github.com/HubuKG/MANoR}. 
\end{abstract}

\section{Introduction}
Multimodal Knowledge Graphs (MMKGs) \cite{1} extend conventional knowledge graphs by associating entities with heterogeneous modalities, including images, audio, videos, and textual descriptions. Such information is increasingly important for building AI systems that reason over entities not only as symbolic identifiers, but also as objects embedded in real-world multimodal contexts. By connecting structured relational semantics with unstructured perceptual signals, MMKGs provide a representation substrate for multimodal reasoning, recommendation, and knowledge-enhanced foundation models \cite{2,3,4,6}.

Learning effective entity representations over MMKGs is a central problem for multimodal knowledge graph completion (MMKGC) \cite{ADAMF}, where the model must infer missing triples from both relational structure and multimodal content. Unlike unimodal KGC \cite{10,Distmult}, MMKGC requires the model to handle heterogeneous modalities whose semantic granularity, noise level, and relation-specific usefulness can differ substantially. Therefore, an effective MMKGC model needs to preserve modality-specific semantics, exploit useful cross-modal complementarity, and prevent unreliable signals from dominating entity representations during relational reasoning.

\begin{figure}
    \centering
    \includegraphics[width=0.7\linewidth]{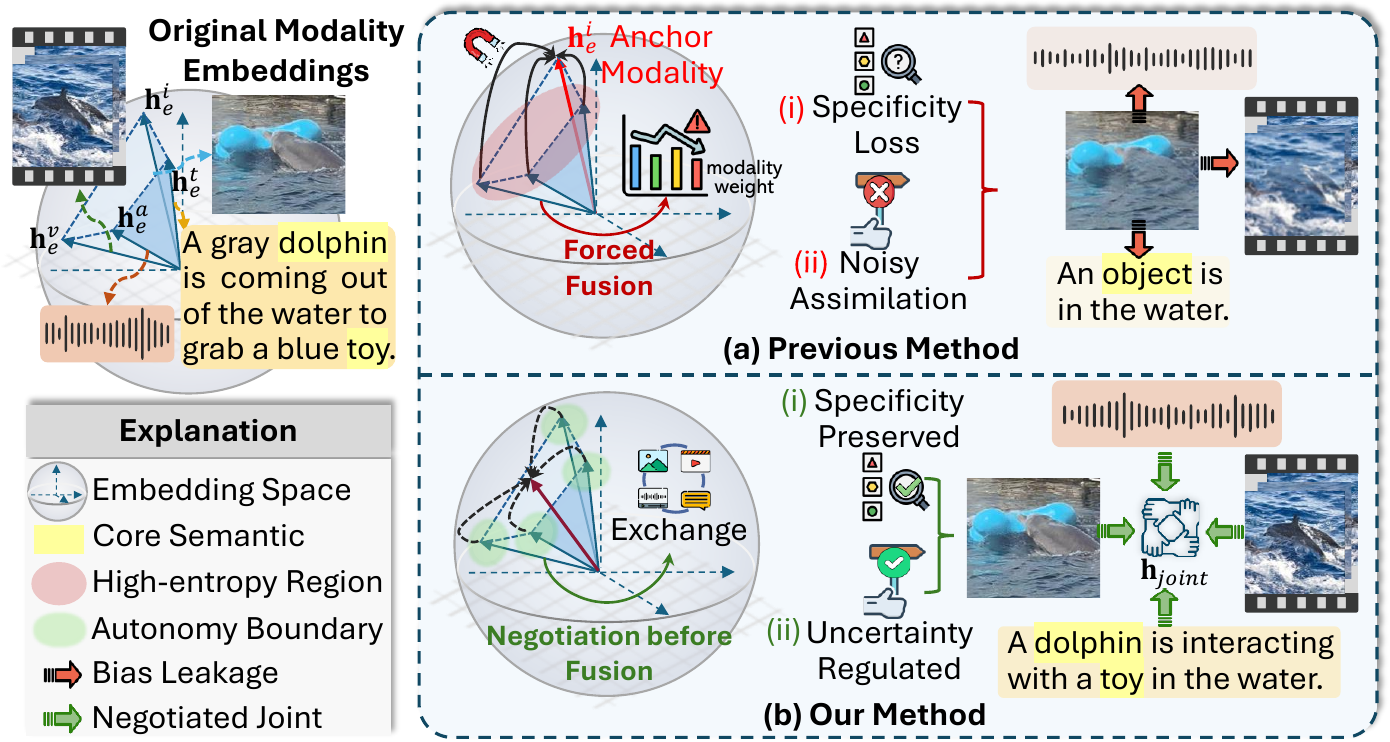}
    \caption{Comparison between (a) traditional fusion strategies may incur specificity loss and noisy assimilation from their rigid alignment, and (b) our proposed negotiation-based-fusion, which maintains a modality autonomy boundary to preserve inherent strengths and calibrate uncertainty.}
    \label{fig1}
\end{figure}

Most existing MMKGC methods \cite{MOCME,MCKGC} address these challenges through attention-based aggregation. While effective in many settings, these mechanisms often impose a rigid alignment among modalities. In practice, such alignment can bias the representation space toward a dominant anchor modality \cite{17}. Once a modality becomes dominant, other modalities are encouraged to adapt to its geometry, even when they contain complementary or more fine-grained information. This tendency is related to the path-of-least-resistance behavior observed in multimodal learning \cite{16,liu2026}, where optimization favors signals that are easier to fit rather than those that are more semantically faithful.

This rigid alignment leads to two recurring failure modes, as illustrated in Figure~\ref{fig1}(a). The first is \textbf{specificity loss}. When non-anchor modalities are forced to match the anchor geometry, their distinctive semantic cues may be diluted. For instance, a detailed textual description such as “\textit{a gray dolphin grabbing a blue toy}” may be reduced to a generic concept such as “\textit{an object}”. The second is \textbf{noisy assimilation}. If the dominant modality is ambiguous or noisy, other modalities may absorb this uncertainty during fusion, producing representations that appear confident but are semantically unreliable. These problems become more severe when modalities differ significantly in noise level or semantic resolution.

To address these challenges, we propose a novel framework called MANoR, which rethinks fusion as a negotiation process rather than rigid geometric alignment, as shown in Figure \ref{fig1}(b). Our key insight is that distinct modalities should be treated as independent agents capable of “agreeing to disagree.” Specifically, we introduce the Modality Autonomous Negotiation (MoA) module, which leverages subspace centroid projection to distill the semantic essence from noisy raw features, and then regulates cross-modal information exchange through attention bootstrapping. This design prevents semantic drift while preserving fine-grained core information. To further enhance robustness against low-quality or noisy data, we devise the Uncertainty Calibrated Fusion (UnCF) module. By modeling aleatoric uncertainty, UnCF acts as a probabilistic filter that dynamically penalizes high-entropy features and calibrates global modality biases. Finally, we integrate these negotiated and calibrated representations into a generative framework to jointly optimize the MMKGC task. Extensive experiments on multiple benchmarks demonstrate the effectiveness of MANoR.

Our contributions are summarized as follows:
\begin{itemize}
\item We propose MANoR, a novel framework that redefines multimodal fusion as an uncertainty-calibrated autonomous negotiation process. This framework enables modalities to preserve their intrinsic semantics while interacting selectively, thereby ensuring effective fusion.
\item We develop two synergistic modules: Modality Autonomous Negotiation and Uncertainty Calibrated Fusion. They jointly combine centroid-guided subspace contraction with dynamic uncertainty
calibration.
\item We conduct extensive experiments on standard benchmarks, demonstrating that MANoR substantially outperforms strong baselines. Specifically, on the KVC16K dataset, it achieves gains of 17.41\% in MRR, 17.72\% in Hits@1, and 10.47\% in Hits@10.
\end{itemize}

\section{Related Work}
\paragraph{Multimodal Knowledge Graph Completion.}

MMKGC aims to infer missing links by jointly exploiting relational structure and heterogeneous modality information \cite{vbkgc}. Early translation-based methods, such as IKRL \cite{IKRL} and TransAE \cite{TransAE}, projected visual and textual features into a unified vector space to constrain structural embeddings. Subsequent works explore more expressive fusion mechanisms to handle complex multimodal interactions, such as IMF \cite{IMF} and AdaMF \cite{ADAMF}. To address the varying importance of modalities across relational contexts, MoMoK \cite{MOMOK} and MoCME \cite{MOCME} introduced mixture-of-experts architectures, allowing different experts to specialize in different relational contexts. Other works focus on more specific challenges in multimodal representation learning. MyGO \cite{MYGO} addresses granularity mismatch by modeling modality information across semantic levels, while APKGC \cite{APKGC} introduces attention penalties to reduce over-confident reliance on noisy modality signals. Recent approaches further explore contrastive learning and unified representation spaces to improve cross-modal alignment \cite{18,21}.

\paragraph{Uncertainty Quantification.}
Uncertainty quantification (UQ) has been widely studied for improving the robustness
of multimodal learning \cite{UQ4,UQ1,UQ3}. Classical Bayesian methods
\cite{bayesian1} and deep ensembles \cite{ensemble} provide principled estimates of
predictive uncertainty, but their computational overhead can be substantial for
large-scale multimodal knowledge graphs. As a result, recent work has increasingly
focused on lightweight uncertainty estimation mechanisms that can guide feature
selection, modality weighting, or dynamic fusion. Variational methods have been
used to estimate modality confidence and adjust fusion weights \cite{23,22}, while
Gaussian uncertainty regressors have been introduced to calibrate latent
distributions in multimodal fusion \cite{cold}. Temporal-invariant learning has
also been explored to reduce redundant or unstable multimodal features \cite{25}.

Unlike methods that use UQ mainly as a regularizer or post-hoc confidence score,
MANoR uses uncertainty to calibrate the fusion process itself, allowing unreliable
feature dimensions and modalities to contribute less during relation-aware fusion.

\section{Methodology}
\label{sec3}
\begin{figure*}
    \centering
    \includegraphics[width=1\textwidth]{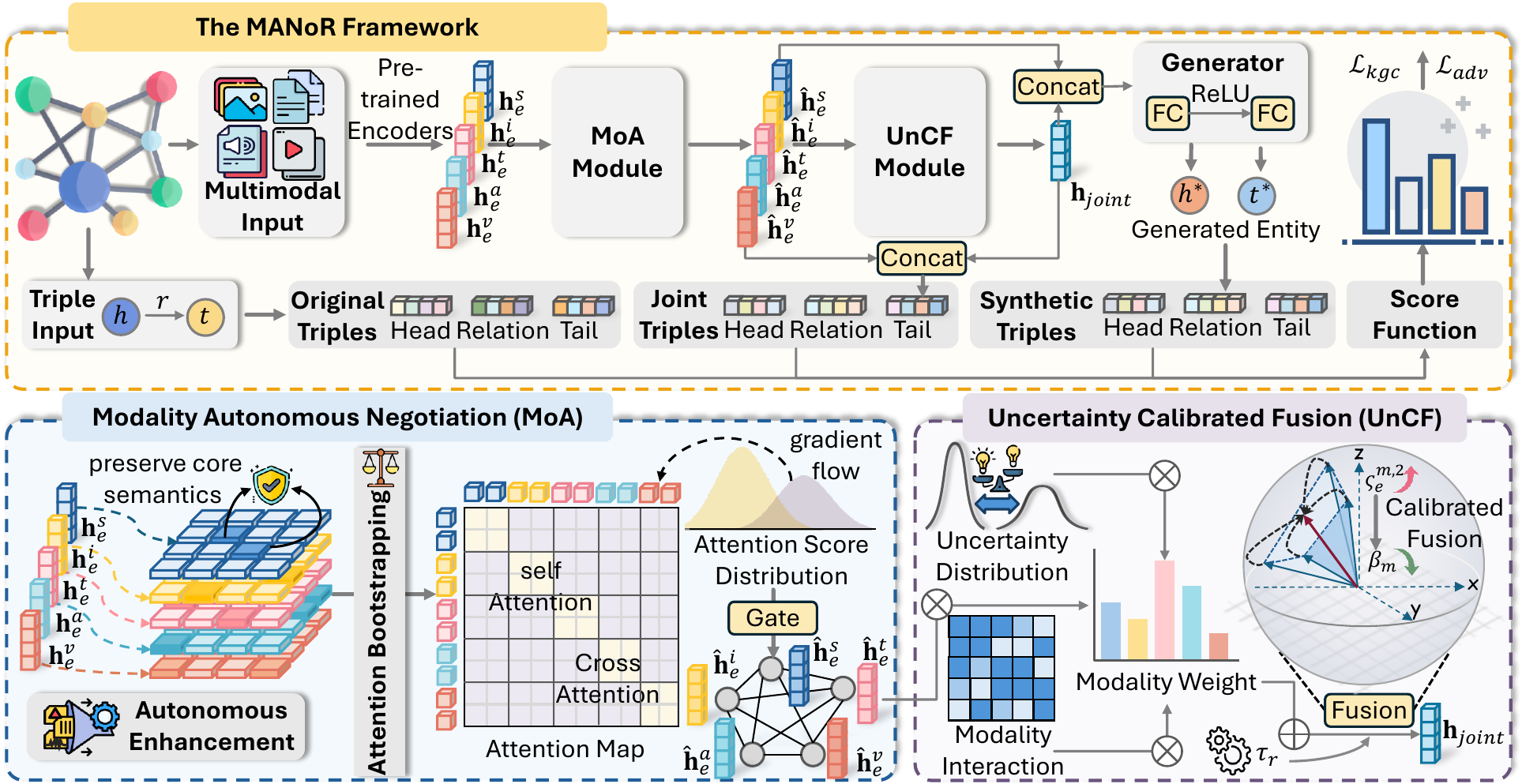}
    \caption{The overview of MANoR. The framework takes multimodal information and triples as inputs, integrates the Modality Autonomous Negotiation module to preserve modality-specific information and support cross-modal interaction, and the Uncertainty Calibrated Fusion module to down-weight noise via uncertainty estimation. Joint embeddings are finally optimized by a generative adversarial scheme for robust MMKGC.}
    \label{fig2}
\end{figure*}

In this section, we propose MANoR, a multimodal representation learning framework illustrated in Figure \ref{fig2}. MANoR combines Modality Autonomous Negotiation for preserving specificity and Uncertainty Calibrated Fusion for mitigating noise propagation. Additionally, a generative adversarial scheme is employed to further strengthen the robustness of the learned representations.

\subsection{Preliminaries}
\paragraph{Problem Formulation.}
A KG is defined as $\mathcal{G}=(\mathcal{E},\mathcal{R},\mathcal{T})$, where $\mathcal{E}$, $\mathcal{R}$, and $\mathcal{T}$ denote sets of entities, relations, and triples $(h,r,t)$, respectively. Furthermore, MMKGs have a modality set denoted as $\mathcal{M}$, encapsulating different modalities (image $I$, text $T$, audio $A$, video $V$) in the MMKGs. For each entity $e\in\mathcal{E}$, its raw data in modality $m\in\mathcal{M}$ is denoted as $Z_m\left(e\right)$. The graph structure itself constitutes an intrinsic modality, whose information is embodied in the triple set $\mathcal{T}$. 

MMKGC aims to learn a scoring function $S\left(h,r,t\right):\mathcal{E}\times\mathcal{R}\times\mathcal{E}\rightarrow\mathbb{R}$ to measure triple plausibility. Training employs negative sampling by corrupting entities in valid triples to form a negative set $\mathcal{T}'$. During inference, the model is evaluated via link prediction \cite{10}, ranking candidates for queries $(?,r,t)$ or $(h,r,?)$ \cite{ADAMF}.
\paragraph{Pre-trained Multimodal Representation.}

Following \cite{native}, we employ pre-trained modality encoders to extract initial
features for each modality. For entity $e$ and modality $m$, the initial
feature is computed as:
\begin{equation}
    \mathbf{e}_m = \frac{1}{| Z_m\left(e\right)|} \sum_{x \in Z_m\left(e\right)} \Phi_m(x),
\end{equation}
where $\Phi_m$ is the corresponding pre-trained encoder, such as BERT \cite{BERT}
for text features. To unify representations across modalities, we project features into a shared semantic space via a lightweight adapter $\mathcal{P}_m$, implemented as a two-layer MLP with ReLU activation: $\mathbf{h}_e^m = \mathcal{P}_m(\mathbf{e}_m) \in \mathbb{R}^{d_e}$.

\subsection{Modality Autonomous Negotiation}
To handle modality heterogeneity without sacrificing semantic specificity, we propose Modality Autonomous Negotiation (MoA). MoA models modalities as independent agents that selectively leverage external context to enhance representations without distorting their internal feature manifolds.

\paragraph{Autonomous Enhancement}
Raw modality features contain both stable semantic factors and unstable variations.
Directly mixing them across modalities may amplify semantic drift, especially when
one modality is noisy or overly dominant \cite{zhixin}. Therefore, MoA stabilizes
each modality within its own principal semantic subspace before cross-modal
interaction. For modality $m$, we estimate a modality-specific basis $U_m$ from the centered
covariance of projected entity features. This basis retains the dominant covariance
directions and defines the subspace where the main modality-specific variations are
represented:
\begin{equation}
U_m=\arg\max_{U^\top U=I}\mathrm{Tr}(U^\top C_mU),
\end{equation}
where $C_m$ denotes the empirical covariance matrix. For simplicity, $\mathcal{E}$
denotes the training entities with available modality-$m$ features in this
subsection. The projected semantic factor and its modality centroid are:
\begin{equation}
\mathbf{s}_e^m=U_m^\top \mathbf{h}_e^m,\qquad
\boldsymbol{\mu}^m=\frac{1}{|\mathcal{E}|}
\sum_{e\in\mathcal{E}}\mathbf{s}_e^m.
\end{equation}
MoA then refines each entity by contracting its centroid-relative deviation in the
principal subspace:
\begin{equation}
\tilde{\mathbf{h}}_e^m
=
\mathbf{h}_e^m-\lambda_m U_m(\mathbf{s}_e^m-\boldsymbol{\mu}^m),
\end{equation}
where $\tilde{\mathbf{h}}_e^m$ is the enhanced representation, and
$U_m(\mathbf{s}_e^m-\boldsymbol{\mu}^m)$ maps the centroid-relative deviation back
to the original feature space. The coefficient
$\lambda_m=\sigma(\theta_m)\in(0,1)$ is a learnable modality-specific contraction
coefficient, where $\theta_m\in\mathbb{R}$ is unconstrained and $\sigma(\cdot)$ is
the sigmoid function. Larger $\lambda_m$ yields stronger contraction toward the
modality centroid.

Since $U_m^\top U_m=I$, the refined projected factor satisfies
$\tilde{\mathbf{s}}_e^m-\boldsymbol{\mu}^m
=(1-\lambda_m)(\mathbf{s}_e^m-\boldsymbol{\mu}^m)$, where
$\tilde{\mathbf{s}}_e^m=U_m^\top\tilde{\mathbf{h}}_e^m$. Thus, the update reduces
centroid-relative dispersion only within the selected subspace, while leaving the
orthogonal residual $(I-U_mU_m^\top)\mathbf{h}_e^m$ unchanged. This controlled
contraction stabilizes modality-specific representations without collapsing
entity-level discriminative information.

\paragraph{Attention Bootstrapping}
After autonomous enhancement, each modality obtains a stabilized representation
$\tilde{\mathbf{h}}_e^m$ but remains isolated from other modalities. The goal of attention bootstrapping is not immediate fusion, but to probe potential cross-modal relevance while maintaining modality autonomy. For entity $e$, we define modality-wise latent
probes via linear transformations:
\begin{equation}
\mathbf{q}_e^m = \mathbf{W}_q^m \tilde{\mathbf{h}}_e^m, \quad
\mathbf{k}_e^m = \mathbf{W}_k^m \tilde{\mathbf{h}}_e^m, \quad
\mathbf{v}_e^m = \mathbf{W}_v^m \tilde{\mathbf{h}}_e^m,
\end{equation}
where $\mathbf{W}_q^m, \mathbf{W}_k^m \in \mathbb{R}^{d_k \times d_e}$ are learnable projection matrices, while $\mathbf{W}_v^m \in \mathbb{R}^{d_e \times d_e}$ preserves the semantic dimension and $d_k$ is the attention head dimension. We then define a soft relevance matrix capturing cross-modal affinity:
\begin{equation}
\alpha_e^{m \rightarrow n}=
\frac{
\exp \left( \langle \mathbf{q}_e^m , \mathbf{k}_e^n \rangle / \sqrt{d_k} \right)
}{
\sum_{n' \in \mathcal{M}}
\exp \left( \langle \mathbf{q}_e^m , \mathbf{k}_e^{n'} \rangle / \sqrt{d_k} \right)
}.
\end{equation}
This mechanism allows each modality to identify complementary evidence from other modalities. Rather than directly overwriting modality features, we interpret this attention distribution as a bootstrap signal. The bootstrapped response is computed as:
\begin{equation}
\mathbf{b}_e^m=
\sum_{n \in \mathcal{M}}
\alpha_e^{m \rightarrow n}\mathbf{v}_e^n.
\end{equation}

\paragraph{Negotiated Update}
Directly injecting the bootstrapped response may dilute modality-specific semantics.
We therefore use a dimension-wise negotiation gate
$\boldsymbol{\gamma}_e^m$ to control the incorporation of cross-modal information:
\begin{equation}
\boldsymbol{\gamma}_e^m
=
\sigma
\Big(
\mathbf{W}_g^m
\big[
\tilde{\mathbf{h}}_e^m \,\|\, \mathbf{b}_e^m
\big]
+
\mathbf{b}_g^m
\Big),
\end{equation}
where $\sigma(\cdot)$ is the sigmoid function, $[\cdot \,\|\, \cdot]$ denotes concatenation and $\mathbf{W}_g^m \in \mathbb{R}^{d_e \times 2d_e}$ is the gating projection. The updated representation is computed as:
\begin{equation}
\hat{\mathbf{h}}_e^m
=
\boldsymbol{\gamma}_e^m \odot \mathbf{b}_e^m
+
(1 - \boldsymbol{\gamma}_e^m) \odot \tilde{\mathbf{h}}_e^m,
\end{equation}
where $\odot$ denotes the Hadamard product. Smaller values of
$\boldsymbol{\gamma}_e^m$ preserve the autonomous representation, whereas larger
values incorporate more cross-modal information.

\subsection{Uncertainty Calibrated Fusion}

Although negotiated representations incorporate complementary cross-modal evidence,
their reliability is still uneven across entities, feature dimensions, and modalities.
For example, visual signals can be informative for some relations but
ambiguous for others. To reduce such noise propagation,
we propose the Uncertainty Calibrated Fusion (UnCF) module, which estimates
aleatoric uncertainty for each negotiated modality representation and uses it to
calibrate relation-aware modality fusion.

\paragraph{Aleatoric Uncertainty Estimation}

After MoA, each modality obtains a negotiated representation
$\hat{\mathbf{h}}_e^m$ that preserves modality-specific semantics while incorporating
selective cross-modal evidence. We associate this representation with a diagonal
Gaussian distribution:
\begin{equation}
\mathbf{z}_e^m \sim \mathcal{N}(\hat{\mathbf{h}}_e^m, \operatorname{diag}(\boldsymbol{\varsigma}_e^{m,2})),
\end{equation}
where $\mathbf{z}_e^m$ denotes the latent probabilistic embedding. The diagonal variance is estimated as:
\begin{equation}
\boldsymbol{\varsigma}_e^{m,2}
= \operatorname{Softplus}(\mathbf{W}_{\varsigma}^m \hat{\mathbf{h}}_e^m + \mathbf{b}_{\varsigma}^m) + \epsilon,
\end{equation}
where $\boldsymbol{\varsigma}_{e}^{m,2}$ captures dimension-wise
aleatoric uncertainty, $\mathbf{W}_{\varsigma}^m$
and $\mathbf{b}_{\varsigma}^m$ are learnable parameters, and $\epsilon$ is a small constant for numerical stability. A larger variance in a specific dimension
corresponds to higher marginal uncertainty of that semantic dimension. Such dimensions
are therefore assigned lower reliability in the subsequent fusion stage.

\paragraph{Calibrated Fusion}
To mitigate noise propagation, UnCF first constructs a reliability-aware scoring
feature by attenuating uncertain dimensions:
\begin{equation}
\tilde{\mathbf{u}}_e^m
=
\hat{\mathbf{h}}_e^m
\odot
\frac{1}{1+\boldsymbol{\varsigma}_e^{m,2}},
\end{equation}
where $\tilde{\mathbf{u}}_e^m$ is the calibrated feature used for modality scoring.
The attenuation is applied element-wise, so dimensions with larger predicted
variance contribute less to the fusion weight. Besides dimension-level uncertainty, each modality may exhibit a persistent reliability bias during training. We summarize this bias by an exponential moving average:
\begin{equation}
y_e^m
=
\frac{1}{d_e}
\sum_{j=1}^{d_e}
\log\left(1+\boldsymbol{\varsigma}_{e,j}^{m,2}\right),
\quad
\bar a_m^{(t)}
=
\eta \bar a_m^{(t-1)}
+
(1-\eta)
\frac{1}{|\mathcal{B}|}
\sum_{e\in\mathcal{B}}y_e^m ,
\end{equation}
where $y_e^m$ summarizes the average uncertainty, $t$ is the training iteration, $\eta$ is the momentum coefficient, and $\mathcal{B}$ is the current mini-batch. 
For relation-aware fusion, the calibrated modality weight is computed as:
\begin{equation}
\beta_m(e,r)
=
\frac{
\exp\left(
\frac{\alpha_r}{1+\bar a_m^{(t)}}
\mathbf{V}_m^\top \tilde{\mathbf{u}}_e^m
\right)
}{
\sum_{k\in\mathcal{M}}
\exp\left(
\frac{\alpha_r}{1+\bar a_k^{(t)}}
\mathbf{V}_k^\top \tilde{\mathbf{u}}_e^k
\right)
},
\end{equation}
where $\mathbf{V}_m$ is a learnable modality-specific scoring vector, and
$\alpha_r=\operatorname{Softplus}(\tau_r)+\epsilon$ is a relation-specific inverse
temperature with learnable scalar $\tau_r\in\mathbb{R}$. The factor
$1+\bar a_m^{(t)}$ down-weights globally unreliable modalities, while $\alpha_r$
controls the sharpness of modality selection for relation $r$.
The final joint representation is obtained by aggregating the negotiated modality
representations:
\begin{equation}
\mathbf{h}_{joint} = \sum_{m \in \mathcal{M}} \beta_m(e, r) \hat{\mathbf{h}}_e^m.
\end{equation}
Thus, UnCF combines dimension-level uncertainty attenuation, modality-level reliability
normalization, and relation-aware evidence selection in a unified fusion process.
\subsection{Overall Training Objective}

To jointly optimize structural reasoning and multimodal representation learning, we
train MANoR with a link prediction objective and an adversarial generation objective.
Given the joint entity embeddings $\mathbf{h}_{joint}$ and $\mathbf{t}_{joint}$,
we use the RotatE scoring function \cite{rotate} to measure the plausibility of a
triple $(h,r,t)$. The scoring function is defined as:
\begin{equation}
F(h, r, t) = -\|\mathbf{h}_{joint} \circ \mathbf{r} - \mathbf{t}_{joint}\|,
\end{equation}
where $\circ$ is the rotation operator. The link prediction loss is optimized via self-adversarial negative sampling:
\begin{equation}
\mathcal{L}_{kgc} = - \log \sigma(\delta + F(h, r, t))-\sum_{i=1}^{K} p(h'_i, r, t'_i) \log \sigma(-F(h'_i, r, t'_i) - \delta),
\end{equation}
where $\delta$ is the margin and $K$ is the number of negative samples. The negative
set $\mathcal{T}'=\{(h'_i,r,t'_i)\}_{i=1}^{K}$ is generated by corrupting the head
or tail entity of a positive triple $(h,r,t)$, and the sampling weights
$p(h'_i,r,t'_i)$ follow the self-adversarial strategy of RotatE \cite{rotate}.

Following NativE \cite{native}, we further introduce a two-layer MLP generator
$\mathcal{G}$ to synthesize hard negative entities. The generator maps real entity
embeddings together with random noise into the latent space. Given a triple $(h, r, t)$, $\mathcal{G}$ generates synthetic entities $h^*$ and $t^*$ to construct the set of synthetic triples
$\mathcal{S}(h,r,t)=
\{(h^*,r,t),(h,r,t^*),(h^*,r,t^*)\}$.
The adversarial loss encourages the discriminator to assign higher
plausibility to real triples than to generated synthetic triples:
\begin{equation}
\mathcal{L}_{adv} = \sum_{(h, r, t) \in \mathcal{T}} \left( -F(h, r, t) + \frac{1}{|\mathcal{S}|} \sum_{\substack{(h^*, r, t^*) \in \mathcal{S}(h, r, t)}} F(h^*, r, t^*) \right).
\end{equation}
To stabilize adversarial training, we apply a gradient penalty on synthetic triples:
\begin{equation}
\mathcal{L}_{gp} = \sum_{(h^*, r, t^*) \in \mathcal{S}} \left( \left\|\nabla F(h^*, r, t^*)\right\|_2 - 1 \right)^2.
\end{equation}
The model is optimized via an alternating strategy, where the discriminator minimizes the KGC and adversarial losses, while the generator minimizes the adversarial loss regularized by the gradient penalty:
\begin{equation}
\left\{
\begin{aligned}
\mathcal{L}_{D} &= \mathcal{L}_{kgc} + \lambda_1 \mathcal{L}_{adv}, \\
\mathcal{L}_{G} &= -\mathcal{L}_{adv} + \lambda_2 \mathcal{L}_{gp},
\end{aligned}
\right.
\end{equation}
where $\lambda_1$ and $\lambda_2$ are hyperparameters balancing knowledge retention and adversarial generation.

\begin{table*}[t]
\centering
\small
\setlength{\tabcolsep}{3pt}
\renewcommand{\arraystretch}{0.95}
\resizebox{\textwidth}{!}{
\begin{tabular}{lcccccccccccc}
\toprule
\multirow{2}{*}{\textbf{Method}} &
\multicolumn{3}{c}{\textbf{MKG-W}} &
\multicolumn{3}{c}{\textbf{MKG-Y}} &
\multicolumn{3}{c}{\textbf{TIVA}} &
\multicolumn{3}{c}{\textbf{KVC16K}} \\
\cmidrule(lr){2-4}\cmidrule(lr){5-7}\cmidrule(lr){8-10}\cmidrule(lr){11-13}
& MRR & Hits@1 & Hits@10 & MRR & Hits@1 & Hits@10 & MRR & Hits@1 & Hits@10 & MRR & Hits@1 & Hits@10 \\
\midrule
TransE \cite{10}  & 29.19 & 21.06 & 44.23 & 30.73 & 23.45 & 43.37 & 83.85 & 83.20 & 84.15 & 8.54 & 0.64 & 23.42 \\
DistMult \cite{Distmult}   & 20.99 & 15.93 & 30.86 & 25.04 & 19.33 & 35.95 & 82.27 & 81.15 & 84.22 & 6.37 & 3.03 & 12.61 \\
ComplEx \cite{CompIEX}    & 24.93 & 19.09 & 36.73 & 28.71 & 22.26 & 40.93 & 80.67 & 77.67 & 86.10 & 12.85 & 7.48 & 23.18 \\
RotatE \cite{rotate}     & 33.67 & 26.80 & 46.73 & 34.95 & 29.10 & 45.30 & 84.59 & 83.47 & 86.95 & 14.33 & 8.25 & 26.17 \\
\midrule
IKRL \cite{IKRL}      & 32.36 & 26.11 & 44.07 & 33.22 & 30.37 & 38.26 & 67.71 & 63.72 & 75.67 & 11.11 & 5.42 & 22.39 \\
TransAE \cite{TransAE}   & 30.00 & 21.23 & 44.72 & 28.10 & 25.31 & 33.03 & 79.57 & 74.95 & 88.07 & 10.81 & 5.31 & 21.89 \\
RSME \cite{RSME}     & 29.23 & 23.36 & 40.43 & 34.44 & 31.48 & 39.09 & 40.01 & 30.55 & 51.35 & 12.31 & 7.14 & 22.05 \\
OTKGE \cite{OTKGE}   & 34.36 & 28.85 & 44.88 & 35.51 & 31.97 & 41.38 & 35.28 & 30.45 & 41.98 & 8.77 & 5.01 & 15.55 \\
IMF \cite{IMF}         & 34.50 & 28.77 & 45.44 & 35.79 & 32.95 & 40.63 & 55.46 & 41.87 & 77.57 & 12.01 & 7.42 & 21.01 \\
QEB \cite{tiva}      & 32.38 & 25.47 & 45.32 & 34.37 & 29.49 & 42.32 & 74.25 & 66.10 & 88.35 & 12.06 & 5.57 & 25.01 \\
MMRNS \cite{MMRNS}    & 35.03 & 28.59 & 47.47 & 35.93 & 30.53 & 45.47 & 83.12 & 83.05 & 83.25 & 13.31 & 7.51 & 24.68 \\
AdaMF \cite{ADAMF}    & 34.27 & 27.21 & 47.21 & 38.06 & 33.49 & 45.48 & 89.60 & 88.43 & 91.27 & 15.05 & 8.96 & 27.10 \\
NativE \cite{native}    & 36.58 & 29.56 & 49.39 & 39.04 & 34.79 & 46.25 & 92.27 & 91.40 & 92.85 & 15.76 & 9.23 & 28.55 \\
MoMoK \cite{MOMOK}      & 35.89 & 30.38 & 45.53 & 37.91 & 35.09 & 42.90 & 89.21 & 87.62 & 91.55 & 16.87 & 10.53 & 29.20 \\
MYGO \cite{MYGO}      & 36.10 & 29.78 & 47.75 & 38.44 & 35.01 & 44.19 & - & - & - & - & - & - \\
MCKGC \cite{MCKGC}      & 36.88 & 31.32 & 47.43 & 38.92 & 35.49 & 45.21 & - & - & - & - & - & - \\
DSoM \cite{DSOM}     & \underline{40.90} & \underline{33.56} & \underline{55.28} & 40.02 & 35.67 & \underline{47.71} & - & - & - & - & - & - \\
MoCME \cite{MOCME}     & 37.79 & 30.81 & - & \underline{40.37} & \underline{36.21} & - & \underline{94.58} & \underline{93.32} & \underline{95.61} & \underline{18.32} & \underline{11.23} & \underline{30.76} \\
\rowcolor{gray!17}
\midrule
MANoR (Ours)                          & \textbf{44.05} & \textbf{37.20} & \textbf{60.54} & \textbf{45.13} & \textbf{40.79} & \textbf{53.26} & \textbf{96.03} & \textbf{95.42} & \textbf{97.56} & \textbf{21.51} & \textbf{13.22} & \textbf{33.98} \\
\rowcolor{gray!17}
Improvement $\uparrow$        & \impor{\scalebox{0.85}{7.70\%}} & \impor{\scalebox{0.85}{10.85\%}} & \impor{\scalebox{0.85}{9.52\%}} & \impor{\scalebox{0.85}{11.79\%}} & \impor{\scalebox{0.85}{12.65\%}} & \impor{\scalebox{0.85}{11.63\%}} & \impor{\scalebox{0.85}{1.53\%}} & \impor{\scalebox{0.85}{2.25\%}} & \impor{\scalebox{0.85}{2.04\%}} & \impor{\scalebox{0.85}{17.41\%}} & \impor{\scalebox{0.85}{17.72\%}} & \impor{\scalebox{0.85}{10.47\%}} \\
\bottomrule
\end{tabular}}
\caption{Link prediction results on four benchmark datasets. The best results are highlighted in \textbf{bold}, while the second-best results are \underline{underlined}.}
\label{tab1}
\end{table*}

\section{Experiments}
\label{sec4}
\subsection{Experimental Setup}
\label{sec4.1}
\paragraph{Datasets and Baselines.}
To evaluate the generalization capability and robustness of MANoR, we conduct
experiments on four benchmark datasets: MKG-W \cite{MMRNS}, MKG-Y \cite{MMRNS}, TIVA \cite{tiva}, and KVC16K \cite{kvc16k}. Detailed dataset statistics are provided in Appendix~\ref{app:dataset_statistics}, Table~\ref{tab:dataset_statistics}. We compare MANoR with strong baselines from two categories. 
(i) \textbf{Unimodal KGC Methods}: TransE \cite{10}, DistMult \cite{Distmult}, ComplEx \cite{CompIEX}, and RotatE \cite{rotate}. 
(ii) \textbf{Multimodal KGC Methods}: IKRL \cite{IKRL}, TransAE \cite{TransAE}, RSME \cite{RSME}, OTKGE \cite{OTKGE}, IMF \cite{IMF}, QEB \cite{tiva}, MMRNS \cite{MMRNS}, AdaMF \cite{ADAMF}, NativE \cite{native}, MoMoK \cite{MOMOK}, MYGO \cite{MYGO}, MCKGC \cite{MCKGC}, DSoM \cite{DSOM} and MoCME \cite{MOCME}. 

\paragraph{Evaluation Metrics.}
Following standard protocols in knowledge graph completion \cite{rotate}, we evaluate the performance of our method using two widely adopted metrics: Mean Reciprocal Rank (MRR) and Hits@K (K=1, 10). To ensure fair evaluation, we adopt the filtered setting \cite{10} by removing all valid triples from the candidate space before ranking.

\subsection{Overall Performance}
Table~\ref{tab1} reports the link prediction results on four benchmark datasets.
MANoR achieves the best performance on all reported metrics. Compared with the
strongest baseline on each dataset, it improves MRR by 7.70\% on MKG-W, 11.79\%
on MKG-Y, 1.53\% on TIVA, and 17.41\% on KVC16K. The gains cover both
two-modality and four-modality datasets, suggesting that MANoR generalizes across
different modality configurations. On KVC16K,
where image, text, audio, and video features are all available, MANoR obtains the
largest relative gain, improving MRR from 18.32 to 21.51 and Hits@1 from 11.23 to
13.22. This suggests that when multimodal information is richer but also more
heterogeneous, directly aggregating modalities is less sufficient, and calibrated
modality selection becomes more important. On TIVA, where the strongest baseline is already highly competitive, MANoR still improves
all metrics, showing that the gain is not limited to low-performing settings. The consistent Hits@1 improvements further show that
MANoR is effective in improving top-ranked predictions, which are especially
important in practical prediction scenarios.

\subsection{Modality Specificity Analysis}

\begin{figure}[t]
    \noindent
    \begin{minipage}[t]{0.48\linewidth}
        \centering
        \includegraphics[width=\linewidth]{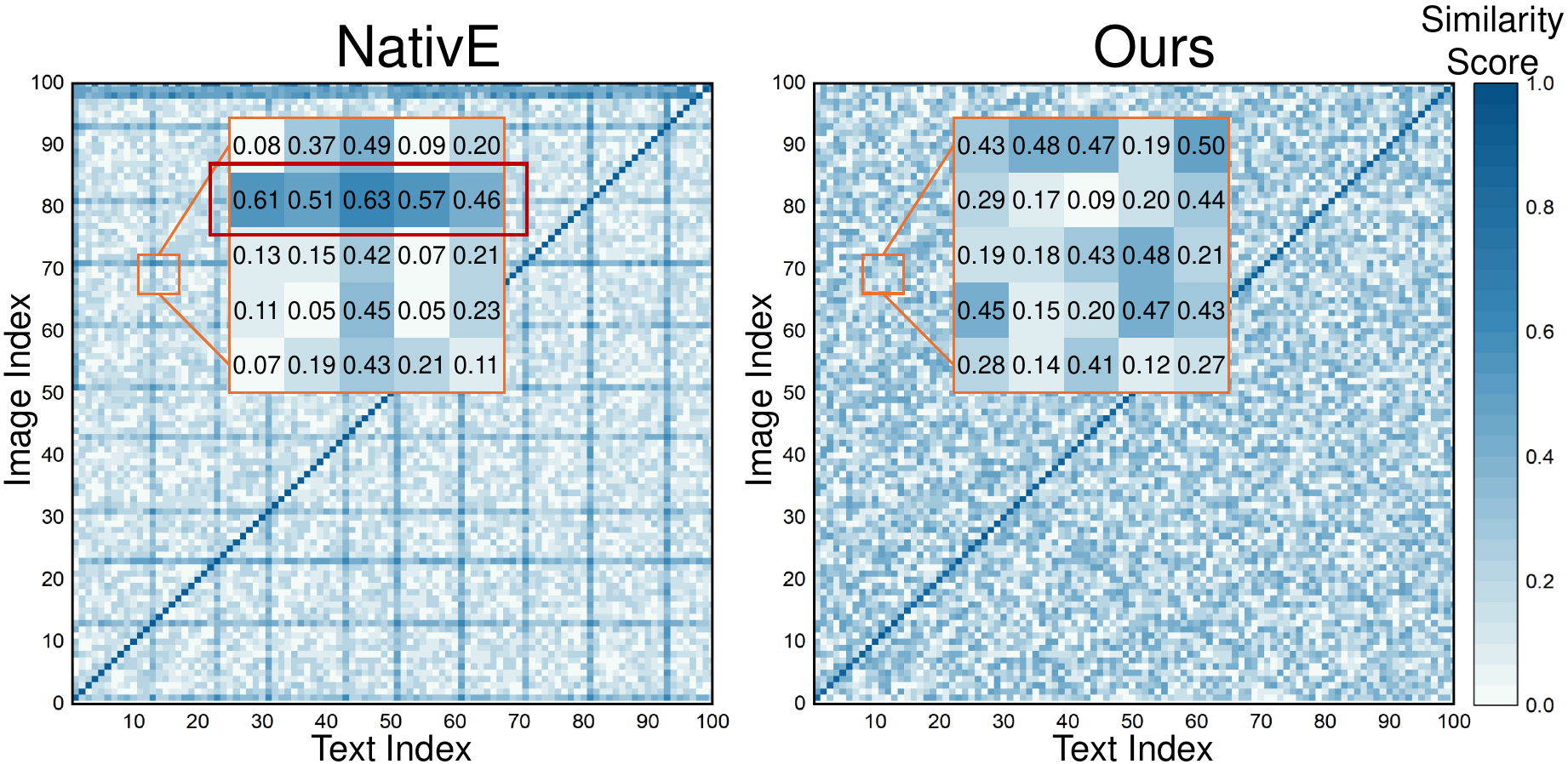}
        \caption{Cross-modal interaction matrices on MKG-Y under image-text matching.}
        \label{fig3}
    \end{minipage}%
    \hfill
    \begin{minipage}[t]{0.48\linewidth}
        \centering
        \includegraphics[width=\linewidth]{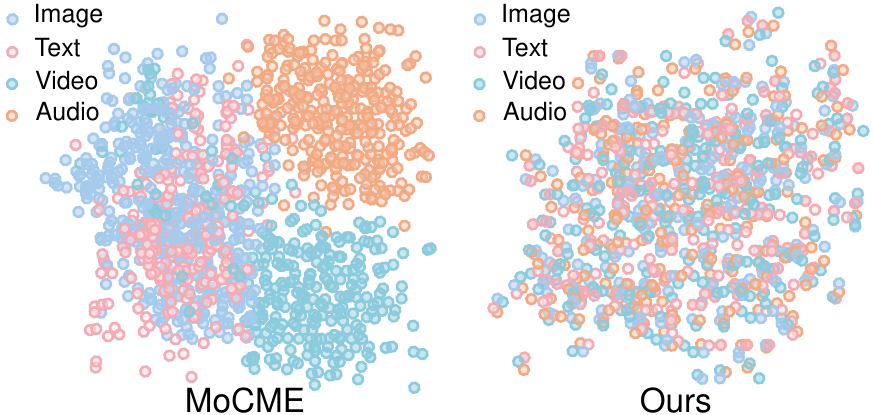}
        \caption{t-SNE visualization of learned modality distributions after fusion.}
        \label{fig4}
    \end{minipage}
\end{figure}

We visualize image-text interaction matrices on MKG-Y to examine whether fusion
preserves modality-specific correspondence. This
analysis directly targets the main failure mode discussed in Figure~\ref{fig1}: if a fusion method collapses different modalities toward a generic anchor, then multiple
non-matching samples may obtain high cross-modal similarity, producing horizontal or vertical striations in the interaction matrix. As shown in Figure~\ref{fig3}, NativE exhibits such off-diagonal striations. This indicates that some embeddings are assigned high
similarity to multiple non-matching samples, suggesting weaker pairwise
discrimination between image and text features. In contrast,
MANoR yields a more diagonally concentrated interaction matrix, where matched image-text pairs
obtain higher similarity and non-matching pairs are less correlated. This pattern indicates that MANoR preserves more discriminative image-text
correspondence, rather than simply increasing global cross-modal similarity. By mitigating early homogenization, MANoR maintains cross-modal discriminability and establishes a more reliable foundation for MMKGC.

\subsection{Representation Visualization}
To further examine how different methods organize multimodal representations, we
visualize the learned feature distributions in Figure~\ref{fig4}. The baseline MoCME exhibits a distinct modality gap. While visual and textual features partially align, audio
and video features remain relatively isolated. This suggests that the fused space
still retains strong modality-dependent separation, which may limit effective
information transfer across modalities. Conversely, MANoR yields a more connected distribution where representations from all modalities are effectively interwoven. This demonstrates that our model successfully abstracts away from the heterogeneity of input signals. MANoR reduces unnecessary modality separation while ensuring that the learned representations are driven by intrinsic semantic content rather than modality-specific variations. Consequently, even highly disparate modalities are effectively integrated into the joint representation.

\subsection{Ablation Study}
\begin{figure}[!t]
\centering

\begin{minipage}[t]{0.5\linewidth}
\vspace{0pt}
\centering
\captionof{table}{The ablation study results on KVC16K. We divide experiments into four groups to isolate the impact of modalities (G1), MoA (G2), UnCF (G3), and the adversarial training strategy (G4). “$\Delta$ (\%)” denotes the relative MRR drop compared
with the full model.}
\label{tab2}

\scriptsize
\setlength{\tabcolsep}{1.5pt}
\renewcommand{\arraystretch}{0.95}

\begin{tabular}{@{}clcccc@{}}
\toprule
Group & Setting & MRR & $\Delta$ (\%) & Hits@1 & Hits@10 \\
\midrule

\multirow{4}{*}{G1}
 & w/o image modality & 19.56 & \drop{9.07} & 10.51 & 30.34 \\
 & w/o text modality  & 19.12 & \drop{11.11} & 11.37 & 31.96 \\
 & w/o audio modality & 20.68 & \drop{3.86} & 13.02 & 33.21 \\
 & w/o video modality & 20.13 & \drop{6.42} & 12.25 & 32.65 \\
\midrule

\multirow{3}{*}{G2}
 & w/o MoA & 18.97 & \drop{11.81} & 11.45 & 31.30 \\
 & w/o $\boldsymbol{\mu}^{m}$ & 19.78 & \drop{8.04} & 11.98 & 33.25 \\
 & w/ Residual as $\boldsymbol{\gamma}_{e}^{m}$
 & 20.35 & \drop{5.39} & 12.57 & 32.91 \\
\midrule

\multirow{2}{*}{G3}
 & w/ Attn. as UnCF & 19.92 & \drop{7.39} & 11.56 & 33.83 \\
 & w/o $\bar{a}_{m}$ & 20.54 & \drop{4.51} & 12.69 & 33.77 \\
\midrule

G4
 & w/o Adv. Training & 21.17 & \drop{1.58} & 13.10 & 33.90 \\
\midrule

\rowcolor{gray!17}
\multicolumn{2}{c}{\textbf{Full MANoR Model}}
 & \multicolumn{2}{c}{\textbf{21.51}}
 & \textbf{13.22}
 & \textbf{33.98} \\
\bottomrule
\end{tabular}
\end{minipage}
\hfill
\begin{minipage}[t]{0.48\linewidth}
\vspace{0pt}
\centering
\includegraphics[width=\linewidth]{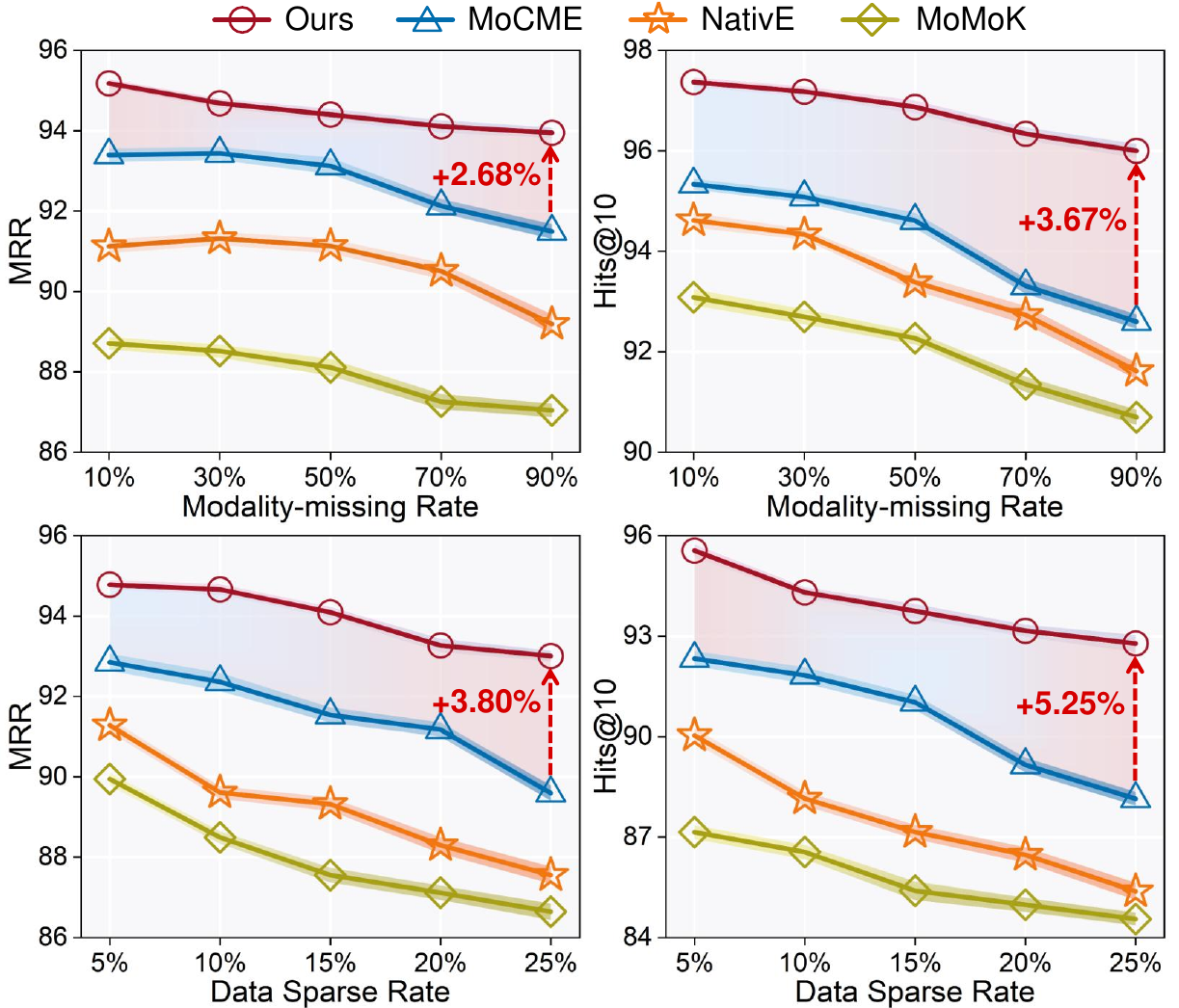}
\captionof{figure}{Performance comparison on TIVA under modality missing
and data sparsity scenarios.}
\label{fig5}
\end{minipage}

\end{figure}

Table~\ref{tab2} reports the ablation results on KVC16K. In G1, removing any
single modality reduces performance, indicating that MANoR does not rely on a single dominant modality but benefits from complementary multimodal information.  In G2, removing MoA causes the largest degradation, with MRR dropping
from 21.51 to 18.97, showing that direct fusion without
autonomous negotiation substantially weakens the learned representation. Removing $\boldsymbol{\mu}^m$ also decreases MRR to 19.78, indicating that the centroid-guided subspace operation is not
redundant. Replacing the negotiation gate with a residual update further reduces performance. This comparison is important because both variants still allow cross-modal
information flow, but the residual variant lacks selective control over what is
incorporated. The gap therefore points to the necessity of controlled, rather than
unconditional, cross-modal updates. In G3, replacing UnCF with standard attention reduces MRR to 19.92,
while removing $\bar{a}_m$ reduces MRR to 20.54. These results show that attention alone can model feature similarity, but it does not explicitly account
for unreliable dimensions or persistently noisy modalities. G4 shows that removing adversarial training leads to only a moderate decline.

\subsection{Complex Scenarios Analysis}
To evaluate robustness under incomplete and limited multimodal information, we test
MANoR under two challenging settings: modality missing and data sparsity. The
results are summarized in Figure~\ref{fig5}.

The top row of Figure~\ref{fig5} shows the results under different modality
missing rates. As the missing rate escalates from 10\% to 90\%, baselines suffer catastrophic degradation, betraying their structural dependency on a dominant anchor. MANoR consistently degrades more slowly than the
baselines and outperforms the strongest baseline by 2.68\% in MRR at the extreme 90\% setting. This suggests that MANoR reduces structural dependency on data completeness by fostering modality autonomy, enabling the reconstruction of robust semantics exclusively from the available evidence.

The bottom row of Figure \ref{fig5} highlights the model's superiority in low-resource settings. As sparsity levels increase, the performance gap widens, with MANoR achieving a 5.25\% gain in Hits@10 at 25\% sparsity. Unlike baselines that overfit to sparse noise, MANoR dynamically down-weights uncertain signals. This prevents the model from memorizing spurious patterns, maintaining robust reasoning capabilities even in data-scarce environments.

\subsection{Case Study}

\noindent
\begin{minipage}{\linewidth}
  \centering
  \includegraphics[width=\linewidth]{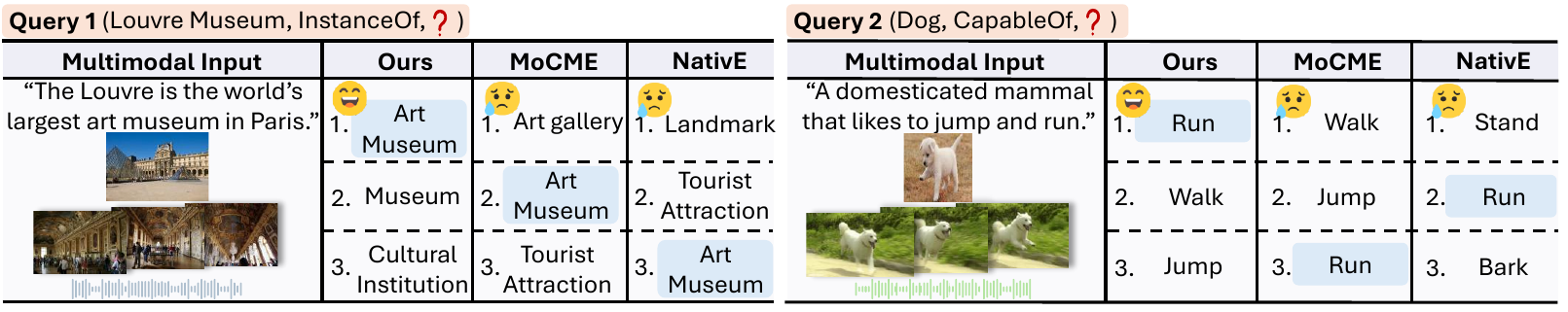}
  \captionof{figure}{Qualitative comparison of Top-3 predictions under
  multimodal inputs.}
  \label{fig6}
\end{minipage}

Figure \ref{fig6} presents a qualitative analysis of real cases from the TIVA dataset. In the first case, NativE biases towards visually salient but semantically generic concepts (e.g., \textit{Landmark} or \textit{Tourist Attraction}), while MoCME struggles to discriminate \textit{Art Museum} from the closely related \textit{Art Gallery}, reflecting semantic over-smoothing caused by forced fusion strategies. However, our method accurately identifies \textit{Art Museum} as the top prediction, followed by semantically coherent candidates, demonstrating its capacity to preserve fine-grained semantic distinctions. Similarly, for the query \textit{(Dog, CapableOf, ?)}, our method prioritizes the specific action \textit{Run}, whereas baselines output generic or static concepts like  \textit{Walk} or \textit{Stand}. These cases suggest that explicitly modeling modality autonomy and uncertainty is essential for achieving precise and semantically faithful multimodal learning.

\section{Conclusion}
\label{sec5}
In this paper, we present MANoR, a multimodal representation learning framework that formulates
MMKGC fusion as uncertainty-calibrated autonomous negotiation. MANoR stabilizes modality-specific
representations before cross-modal interaction and uses estimated reliability to calibrate
the subsequent integration process. Experiments on four MMKGC benchmarks show consistent improvements over strong baselines, including settings with missing modalities and sparse training data. A limitation is that MANoR is mainly
evaluated on MMKGC benchmarks. Future work will investigate
the extension of this approach to broader multimodal reasoning and foundation model
settings, preserving intrinsic modal semantics against text-centric homogenization
to ground linguistic reasoning.

\begin{ack}
\paragraph{Funding.}
This work was supported in part by the National Natural Science Foundation of China (No. 62677024, 62677025, 62377009), the Hubei Provincial Key Research and Development Program (No. 2026BCB079), and the Open Fund of Hubei Key Laboratory of Big Data Intelligent Analysis and Application, Hubei University (No. 2025BDIAA01).
\paragraph{Competing Interests.}
The authors declare no competing interests.
\end{ack}

\bibliographystyle{plainnat}
\bibliography{references}

\newpage

\appendix
\section{Additional Experimental Details}
\label{AppendixA}
\subsection{Dataset Statistics}
\label{app:dataset_statistics}
We provide the detailed statistics of the datasets used in our experiments in
Table~\ref{tab:dataset_statistics}. MKG-W and MKG-Y contain image and text
features, while TIVA and KVC16K provide richer multimodal information including
image, text, audio, and video features. These datasets cover different relation
scales, modality combinations, and feature dimensions, allowing us to evaluate the
generalization ability of MANoR under both two-modality and four-modality MMKG
settings. In the table, Num. denotes the number of entities with available features
for a given modality, and Dim. denotes the corresponding feature dimension. A dash
indicates that the modality is not available in the dataset.

\begin{table}[h]
\centering
\caption{Statistics of the experimental datasets. Num. and Dim. denote the number
of entities with available modality features and the feature dimension,
respectively.}
\label{tab:dataset_statistics}
\resizebox{\textwidth}{!}{
\begin{tabular}{cccccccccccccc}
\toprule
\multirow{2}{*}{\textbf{Dataset}} 
& \multirow{2}{*}{\textbf{\#Entity}} 
& \multirow{2}{*}{\textbf{\#Relation}} 
& \multirow{2}{*}{\textbf{\#Train}} 
& \multirow{2}{*}{\textbf{\#Valid}} 
& \multirow{2}{*}{\textbf{\#Test}}
& \multicolumn{2}{c}{\textbf{Image}} 
& \multicolumn{2}{c}{\textbf{Text}} 
& \multicolumn{2}{c}{\textbf{Audio}} 
& \multicolumn{2}{c}{\textbf{Video}} \\
\cmidrule(lr){7-8}
\cmidrule(lr){9-10}
\cmidrule(lr){11-12}
\cmidrule(lr){13-14}
& & & & & & Num. & Dim. & Num. & Dim. & Num. & Dim. & Num. & Dim. \\
\midrule
\textbf{MKG-W} \cite{MMRNS}  
& 15000 & 169 & 34196  & 4276  & 4274  
& 14463 & 383  & 14123 & 384 & - & - & - & - \\
\textbf{MKG-Y} \cite{MMRNS}  
& 15000 & 28  & 21310  & 2665  & 2663  
& 14244 & 383  & 12305 & 384 & - & - & - & - \\
\textbf{TIVA} \cite{tiva}  
& 11858 & 16  & 20071  & 2000  & 2000  
& 11636 & 2048 & 11858 & 300 & 2441 & 128 & 10269 & 2048 \\
\textbf{KVC16K} \cite{kvc16k} 
& 16015 & 4   & 180190 & 22523 & 22525 
& 14822 & 768  & 14822 & 768 & 14822 & 768 & 14822 & 768 \\
\bottomrule
\end{tabular}
}
\end{table}

\subsection{Implementation Details}
We implement our method in PyTorch and operate it on an NVIDIA A100 GPU. For all datasets, we set the batch size to 1024, the number of attention heads to 4, and the momentum coefficient $\eta$ to 0.9. We initialize each learnable relation-specific parameter $\tau_r$ to 0.01. We tune key hyperparameters over: learning rate \{$1\mathrm{e}^{-4}$, $1\mathrm{e}^{-5}$\}, embedding dimension $d_e$ \{150, 250\}, margin $\delta$ \{3, 4, 9\}, and $\lambda_1$, $\lambda_2$ \{$1\mathrm{e}^{-2}$, $1\mathrm{e}^{-3}$, $1\mathrm{e}^{-4}$, $1\mathrm{e}^{-5}$\}. We fix the optimizer as Adam.

\begin{wraptable}{r}{0.46\columnwidth}
\vspace{-8pt}
\centering
\scriptsize
\setlength{\tabcolsep}{1.6pt}
\renewcommand{\arraystretch}{1}
\caption{Efficiency comparison on different datasets. We report the training time per epoch (minutes) and GPU memory usage (MB).}
\label{tab4}
\begin{tabular}{@{}clcccc@{}}
\toprule
        \multirow{2}{*}{\textbf{Method}} & \multicolumn{2}{c}{\textbf{TIVA}} & \multicolumn{2}{c}{\textbf{KVC16K}} \\
        \cmidrule(lr){2-3} \cmidrule(lr){4-5}
         & \textbf{Time} & \textbf{Memory} & \textbf{Time} & \textbf{Memory} \\
        \midrule
        MMRNS & 8.51 & 24582 & 15.52 & 25600 \\
        NativE & 1.98 & 9723 & 4.10 & 9983 \\
        MoMoK & 3.50 & 11264 & 8.36 & 12288 \\
        MoCME & 6.55 & 17408 & 12.59 & 21491 \\
        \rowcolor{gray!17} 
        \midrule
        \textbf{Ours} & \textbf{2.17} & \textbf{8920} & \textbf{4.68} & \textbf{9831} \\
\bottomrule
\end{tabular}
\vspace{-8pt}
\end{wraptable}
\subsection{Efficiency Analysis}
\label{A.3}
Despite introducing the MoA and UnCF modules, MANoR maintains high efficiency with minimal computational overhead. Table \ref{tab4} shows that incorporating these components leads to only a marginal increase in training time compared to the lightweight baseline NativE. This slight overhead is an expected consequence of the adversarial training mechanism employed to bolster robustness against modal incompleteness. Crucially, MANoR avoids the heavy computational burden observed in MoCME, training significantly faster while achieving the lowest memory consumption on large-scale datasets. Overall, our framework strikes an optimal balance, delivering state-of-the-art performance with competitive time and space efficiency.

\end{document}